\documentclass[runningheads]{llncs}

\usepackage{eccv}

\usepackage{eccvabbrv}

\usepackage{graphicx}
\usepackage{booktabs}
\usepackage{arydshln}
\usepackage{bbm}
\usepackage{nicematrix}
\usepackage{enumitem}
\usepackage[T1]{fontenc}
\usepackage{textcomp}
\usepackage{placeins}
\usepackage{multibib}
\usepackage{comment}
\newcites{Supp}{Supplementary References}

\usepackage[accsupp]{axessibility}  

\usepackage[pagebackref,breaklinks,colorlinks]{hyperref}
\usepackage{orcidlink}
\def\checkmark{\tikz\fill[scale=0.4](0,.35) -- (.25,0) -- (1,.7) -- (.25,.15) -- cycle;}
\usepackage[dvipsnames]{xcolor}

\begin{document}

\title{Zero-Shot Distillation for Image Encoders:\\ How to Make Effective Use of Synthetic Data} 

\titlerunning{Zero-Shot Distillation for Image Encoders}

\author{Niclas Popp¹$^,$², Jan Hendrik Metzen², Matthias Hein¹}

\authorrunning{N. Popp et al.}

\institute{¹University of Tübingen\\ ²Bosch Center for Artificial Intelligence (BCAI), Robert Bosch GmbH}
\maketitle

\begin{abstract}
Multi-modal foundation models such as CLIP have showcased impressive zero-shot capabilities. However, their applicability in resource-constrained environments is limited due to their large number of parameters and high inference time. While existing approaches have scaled down the entire CLIP architecture, we focus on training smaller variants of the image encoder, which suffices for efficient zero-shot classification. The use of synthetic data has shown promise in distilling representations from larger teachers, resulting in strong few-shot and linear probe performance. However, we find that this approach surprisingly fails in true zero-shot settings when using contrastive losses. We identify the exploitation of spurious features as being responsible for poor generalization between synthetic and real data. However, by using the image feature-based $\mathcal{L}_2$ distillation loss, we mitigate these problems and train students that achieve zero-shot performance which on four domain-specific datasets is on-par with a ViT-B/32 teacher model trained on DataCompXL, while featuring up to 92$\%$ fewer parameters.
\end{abstract}

\section{Introduction}
\textbf{Motivation.}
Image classifiers built on top of large vision(-language) foundation models, such as CLIP \cite{clip} or DINOv2 \cite{dinov2}, have shown impressive zero-shot capabilities across various tasks. However, their extensive parameter count and high inference latency present significant challenges for deployment in resource-constrained edge devices used in driver-assistance systems, automated driving, mobile robotics, or video surveillance. Due to their reduced capacity, smaller models cannot be expected to match the performance of larger ones in arbitrary domains. Additionally, training large-scale foundation models typically involves several millions or billions of images, making it expensive and time-consuming. Together, this motivates the need for smaller domain-specific models, as well as data-efficient training procedures. In this work, we specifically focus on zero-shot image classification, for which only a small-scale image encoder is required. Class-specific text embeddings are fixed and can be precomputed off-device, while only image embeddings are computed on-device. Thus, our goal is to distill smaller drop-in replacements (students) of the CLIP image encoder (teacher) that achieve on-par performance on the specific target domains of interest. In particular, we want to specialize the image encoder student to novel domains for which we only know the relevant classes, but do not have access to actual images, the so called \emph{zero-shot distillation} setting. 

\noindent For zero-shot distillation, domain-specific data can be obtained from ``general-purpose'' generative models, such as large-scale latent diffusion models \cite{sdxl}, by class-aware prompting. However, learning from synthetic images has proven  challenging \cite{fake,syntheticImageNet}: using simple class-specific text prompts, like those used for zero-shot classification by CLIP \cite{clip}, yields low-diversity datasets and poor classifiers, as observed in previous studies \cite{syntheticImageNet, fake}. Furthermore, when training on a combination of natural and such low-diversity synthetic images, the overall accuracy starts to decline as synthetic data outweighs real data \cite{syntheticImageNet}. More advanced methods for diversifying prompts with large-language models \cite{diversify}, together with a large compute budget for synthetic image generation, allow achieving high accuracy in linear probe or few-shot scenarios, indicating strong representation learning capabilities. However, we observe in this work that even with diverse prompting, the performance in zero-shot distillation with actual zero-shot evaluation remains comparatively low. We identify that fine-tuning small CLIP image encoders with contrastive text-image losses leads to models that exploit spurious features in images and because of this exhibits poor generalization between synthetic and real images. Linear probe evaluation involves a final linear layer trained on real-data and is thus not affected by the same issue.

\noindent Besides identifying the issue of spurious feature learning, our main contribution is the (somewhat unexpected) observation that a simple change of the distillation loss can mitigate this problem. Specifically, we find that employing a $\mathcal{L}_2$ loss between student's and teacher's image features  substantially reduces the tendency of the models to exploit spurious features and enhances their generalization capabilities between synthetic and real data. We attribute this to the fact that this loss distills image encoders without any influence of the teacher's text encoder and the potential shortcut learning it might encourage. Through one epoch of pre-training on DataComp medium \cite{datacomp} and subsequent fine-tuning on diverse synthetic datasets generated using diffusion models and prompts from large language models, we achieve superior zero-shot classification performance on four target datasets compared to TinyCLIP \cite{tinyCLIP}, the current state-of-the-art for distilled CLIP models. Furthermore, our approach achieves on-par performance with the teacher model, all without utilizing a single annotated image from the target domain during training. When using a image encoder that features only 11 million trainable parameters, we manage to achieve the zero-shot performance of a ViT-B/32 with 86 million parameters within a margin of 5 percent points and outperform a TinyCLIP model with 90 million trainable parameters on three of four test datasets.

\begin{figure}[tb]
  \centering
  \includegraphics[width=\textwidth]{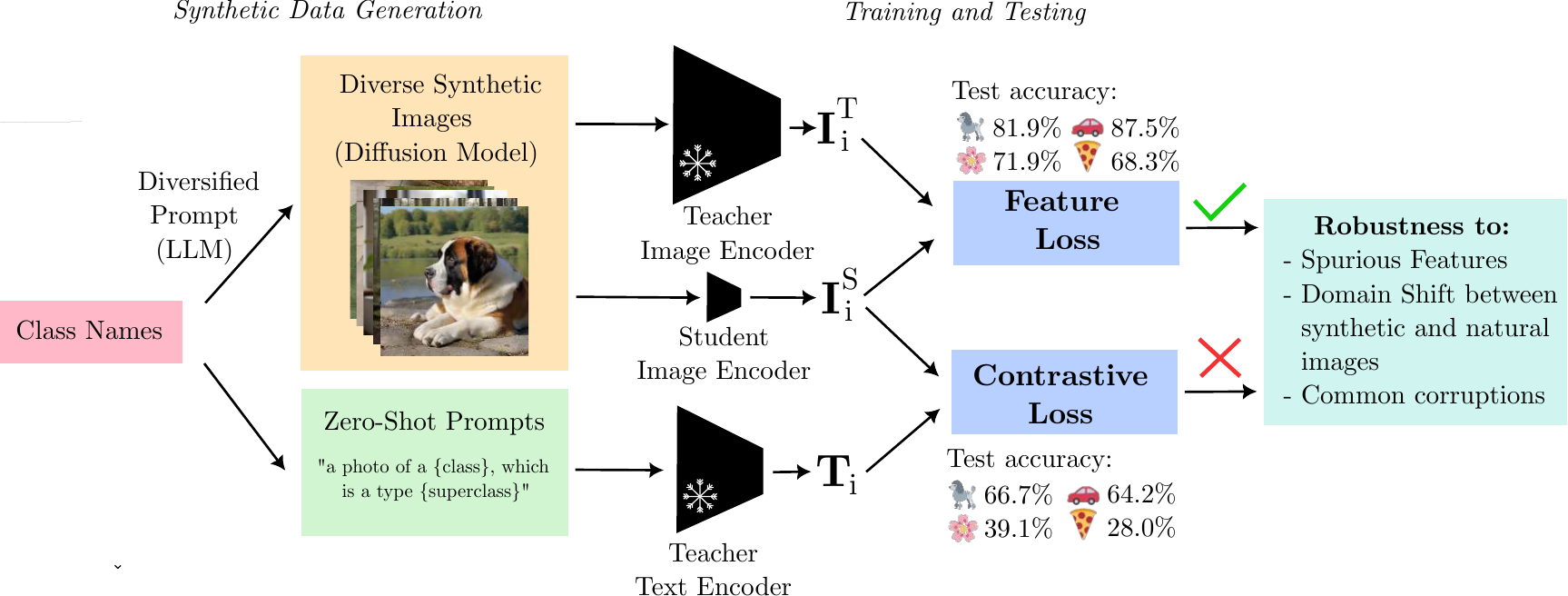}
  \caption{Overview over our zero-shot distillation framework and the observed properties of the contrastive and feature losses. The reported test accuracies are from a fine-tuned TinyViT-11M vision encoder.}
  \label{fig:pipeline}
\end{figure}
\noindent \textbf{Main contributions.}
In this work, we show that small 1-to-1 replacements of the CLIP vision encoder can be efficiently and robustly trained in a zero-shot setting using feature distillation. Our main contributions are the following:
\begin{enumerate}[noitemsep,topsep=0pt,parsep=0pt,partopsep=0pt]
    \item We introduce a unifying framework for zero-shot distillation of image encoders.
    \item We identify failure cases of distillation when using contrastive losses for fine-tuning on small or synthetic datasets. We attribute this to models being prone to learning spurious features and overfitting to the training domain.
    \item We propose feature distillation, which is not susceptible to these spurious features and generalizes better between synthetic and natural images.
    \item Using our framework we manage to distill a ViT-B/32 CLIP vision encoder into student models with up to 93$\%$ fewer parameters that closely match the classification performance of the teacher and surpass existing baselines on the Oxford Pets \cite{pets}, Flowers-102 \cite{flowers}, Stanford Cars \cite{cars} and Food-101 \cite{food} datasets.
\end{enumerate}

\section{Related Work}
\noindent \textbf{Knowledge Distillation of Vision-Language Models.}
Knowledge Distillation \cite{ogdistillation} is a widely used technique for transferring knowledge from larger teacher models to smaller student models. In its vanilla form, the approach involves combining a standard training loss with a distillation loss that considers the output of both the student and teacher models, penalizing discrepancies between the two models. Knowledge distillation has been observed to not only benefit the test accuracy of the student on the target datasets but transfer other favorable properties of the teacher such as domain generalization \cite{domain_distil}. While this approach has been well-established for single-modality tasks including vision \cite{tiny_vit, vision_distill} or language \cite{distilbert,tinybert}, recent works have extended the concept to the multi-modal setting, specifically in the context of vision-language models. CLIP-KD \cite{clipKD} provides an extensive set of experiments comparing various different loss combinations. TinyCLIP \cite{tinyCLIP} proposed an advanced initialization process using weight inheritance from the teacher to the student as well as a multi-stage progressive distillation culminating in models that are only 1/4 the size of a ViT-B/32 CLIP model. MobileCLIP \cite{mobileCLIP} further refined the distillation process by incorporating image augmentation, synthetic captions, and dedicated architectural choices. In contrast to these existing methods, our approach focuses on finding only a one-to-one replacement of the vision encoder while the text encoder remains frozen. Apart from CLIP-specific techniques, unsupervised distillation based purely on images without labels has been identified as a data-efficient alternative to supervised training for vision encoders \cite{L2distillation} and class-incremental generalization \cite{ood_distillation}. We build on this observation by combining unsupervised pre-training with targeted fine-tuning. Despite knowledge distillation being a widely adopted training technique, it has been observed that it does not always work as commonly understood. Even when the student features the same capacity as the the teacher, there can be significant discrepancies in their predictive distributions \cite{does_KD,domain_distil}.

\noindent \textbf{Training and Distillation Using Synthetic Images.}
Recent advancements in generative text-to-image models have sparked a growing interest in utilizing synthetic images for vision applications. Azizi et al. \cite{syntheticImageNet} demonstrated that images from fine-tuned text-to-image models can be combined with real images to enhance the accuracy of classifiers on ImageNet-1k \cite{imagenet}. For text-to-image generation, diffusion models are commonly employed, particularly for knowledge distillation \cite{syntheticDistillation}. However, it was observed that the performance deteriorates when the number of synthetic images surpasses that of real images. Yu et al. \cite{diversify} attributed this decline to the lack of diversity in the used synthetic images. To mitigate this issue, they proposed a strategy to diversify the image generation process by incorporating prompts generated by large language models, thereby enhancing content and style variation. Another approach to diversification in the few-shot setting was presented by Da Costa et al. \cite{diversifiedindomain}, which involved augmentations and low-rank adaptation. By scaling up synthetic datasets, Tian et al.\cite{modelsvsdata} and Hammoud et al. \cite{synthCLIP} demonstrated the feasibility of training vision-language foundation models solely using images from text-to-image models. However, achieving performance on par with or surpassing models trained on real data necessitates the utilization of a large number of synthetic images, on the order of $10^7$ or $10^8$. This not only prolongs the already long training process but also introduces additional computational overhead. Most importantly, the reported results are typically obtained by linear probing or after few-shot training and are not true zero-shot accuracies which we aim to optimize. To mitigate these challenges, we propose a hybrid approach that combines a large-scale dataset of natural images with a smaller set of domain-specific synthetic images. 

\section{Framework for Zero-Shot Distillation}
\label{sec:framework}
Zero-shot distillation refers to the process of transferring  knowledge from a teacher to a student model in a setting where one does not have access to images from the target domain. It is thereby a special case of data-free knowledge distillation which describes the setting where the training data for teacher and the student differ.  Zero-shot distillation specifically focuses on the ability of foundation models as teachers to perform well on unseen data due to their generalization properties. The objective is to transfer this performance to a smaller student model without utilizing any of the unseen data. Therefore, the primary goal is not to address the disparity between the datasets used to train the teacher and student, but rather to extract domain-specific knowledge from the teacher model without having access to the corresponding data. The term zero-shot distillation has been introduced previously \cite{zero_shot_KD}, yet only in the setting for single-modal classifiers that were trained using the cross-entropy loss. In our case, we consider CLIP which is a vision-language model instead of a simple image classifier. In this section, we present a structured framework for zero-shot distillation. Specifically, we discuss the data domains, the training pipeline, the generation of diversified synthetic training data as well as the selection of an appropriate loss function.

\subsection{Data Domain} In the context of (pre-)training for a zero-shot setting, there are currently two core approaches. The first one involves relying on large-scale data such as common crawl datasets \cite{laion,datacomp}. While this approach is feasible for large foundation models, it poses challenges for smaller models as these lack the same level of generalization capabilities due to their smaller capacity. The second approach involves training from scratch using either purely synthetic images \cite{synthCLIP, stablerep, modelsvsdata} or a combination of real and synthetic images \cite{diversify, syntheticImageNet}. Yet by incorporating few-shot learning \cite{synthCLIP,stablerep} on real images or linear probing \cite{synthCLIP,modelsvsdata,stablerep} after training on synthetic data, the reported accuracies are no longer truly zero-shot. We optimize the actual zero-shot performance by adopting a two-stage approach: in the first stage, we pretrain on a large-scale general-purpose dataset consisting of natural images, and subsequently fine-tune using a smaller set of domain-specific synthetic images. This approach allows us to address the limitations of relying solely on generalization or training on synthetic data, and enables us to achieve strong zero-shot performance.


\subsection{Training Pipeline}\label{sec:step1}
In order to shorten training in comparison to training from scratch, Wu et al. \cite{tinyCLIP} have introduced weight inheritance as an initialization scheme for distilling CLIP models. This method has a significant limitation as it can only be applied when the student model shares a similar architecture with the teacher model. Instead of using weight inheritance, we introduce a pre-training step \cite{pretrain} which is not targeted to a specific domain. Pre-training as for large foundation models like the original CLIP \cite{clip} typically requires substantial computational resources due to the use of billions of images. This contradicts the objective of resource-constrained training for small models. He et al. \cite{L2distillation} observed that pre-training can be shortened significantly by using a feature-based loss. For our purpose of training 1-to-1 replacements of CLIP vision encoders, this step has further advantages: by aligning the embeddings of teacher and student, we can mitigate phenomena like the modality gap \cite{modalitygap} where corresponding output vectors are located in different areas of the embedding space. Subsequently, we fine-tune the pre-trained models towards the target domain of interest with the same loss. The only difference between pre-training and fine-tuning is the training data.


\subsection{Data Diversification of Synthetic Images} \label{sec:data_diversification}
\label{sec:diverse_synthetic} In the context of zero-shot learning for image classification, synthetic data generation is based on the class names. However, it has been observed that using only the names to generate images using diffusion models leads to suboptimal performance \cite{fake}. This is primarily due to the lack of diversity in the generated images as well as class ambiguity \cite{diversifiedindomain}. One alternative is to utilize captions from existing real datasets for image generation. This approach is not entirely consistent with a zero-shot setting, as there may not be available captions for all classes. To address this challenge, recent approaches have turned to leveraging large language models (LLMs) to enhance diversity in the prompts. In addition to class names, LLMs are guided by additional inputs for diversification, such as information from a concept bank \cite{synthCLIP} or specific requirements related to contextual and style diversification \cite{diversify}. By incorporating these additional sources of guidance, the generated synthetic data becomes more diverse and aligned with the desired objectives of the target setting. Using the approach from Yu et al. \cite{diversify}, we focus on \textit{contextual dimensions} to achieve diversification. These dimensions are attributes that describe the context of the image such as the background, camera angle, object position, presentation style, and superclasses, all of which are tuned specifically for the target dataset. In contrast to Yu et al. \cite{diversify}, we do not prompt the LLM for each caption separately, but ask for different options for each contextual dimension. This reduces the risk of obtaining similar captions. The final prompt used for the text-to-image model is a comma-separated list of options for these contextual dimensions. Instead of using all possible combinations of options, which would result in a strongly growing number of images given more options, we perform combinatorial testing \cite{comb_test1, comb_test2}.

\subsection{Loss Selection}\label{sec:loss} Knowledge distillation involves combining a training loss $\mathcal{L}_{training}$ with a distillation loss $\mathcal{L}_{distillation}$ \cite{ogdistillation}. The training loss $\mathcal{L}_{training}$ takes into account the image and ground truth labels or captions, while the distillation loss $\mathcal{L}_{distillation}$ is used to align the teacher and student models. Commonly, the overall loss is selected as $\mathcal{L}_{overall} = \mathcal{L}_{training} + \lambda\,\mathcal{L}_{distillation}$, where $\lambda$ is a scaling parameter \cite{KD-survey}. One might assume that using the CLIP loss for $\mathcal{L}_{training}$ would be the most direct approach, as it was used to train the teacher model. However, He et al. observed that pre-training can be substantially sped up by solely employing a feature-based distillation loss without a supervised training loss. In contrast to the CLIP loss, which aims at aligning the text and image embedding of image-caption pairs, the student directly learns from the image features of the teacher without considering the text. More precisely, for every optimization step, we sample a batch of $N$ images $\{\textbf{t}_i,\textbf{i}_i\}_{i=1,...,N}$ as input. Denote the normalized image embedding corresponding to the $i$-th image by $\textbf{I}_i^S$ for the student and $\textbf{I}_i^T$ for the teacher. Based on this, the $\mathcal{L}^2$ feature distillation loss
$
    \mathcal{L}_2^{\textrm{feature}} =  \sum_{i=1}^{N} \| \textbf{I}_i^S - \textbf{I}_i^T \|_2
$
is optimized. For fine-tuning, the CLIP loss function remains the commonly used approach. \cite{finetune}. Yet, it was initially designed to pre-train both the text and image encoder on large-scale vision-language datasets where each image has a distinct caption, for most image classification datasets, each image only possesses a class label instead of a diverse prompts. In this case, we employ the zero-shot captions "a photo of $\{$class name$\}$ which is a type of $\{$superclass$\}$", which were originally introduced for the zero-shot inference of the original CLIP model \cite{clip}, as $\textbf{T}_i$. "Superclass" refers to a general description of the object that can be encountered such as pets, food, cars or similarly. By using these class-specific prompts, several images in a batch may share the same caption which conflicts the goal of decreasing the similarity of images embeddings to the text embeddings of not matching captions in the CLIP loss. An alternative to the CLIP loss is given by the multi-positive contrastive loss introduced in StableRep \cite{stablerep}. To adapt the multi-positive (MP) loss to our setting, we replace the anchor sample by the embedding of a class-specific zero-shot prompt. Denote by $\textbf{Z}_k$ the normalized embedding of the zero-shot prompt for class $k$. The contrastive distribution is given by
$
    q_k =\dfrac{\textrm{exp}(\langle \textbf{I}_i,\textbf{Z}_k \rangle/\tau)}{\sum_{j=1}^{M} \textrm{exp}(\langle \textbf{I}_i,\textbf{Z}_j \rangle/\tau)}
$
and the ground-truth categorical distribution is
$
    p_k = \dfrac{\mathbbm{1}_{l(\textbf{I}_i)=k}}{\sum_{j=1}^{M} \mathbbm{1}_{l(\textbf{I}_i)=j}}
$.
Given $M$ classes, the MP loss is computed as
$
    \mathcal{L}_{\textrm{MP}} = - \sum_{k=1}^{M} p_k\,\textrm{log}\,q_k
$.
We compare contrastive loss, MP loss, feature loss, as well as combinations thereof, on both synthetic and real data in our experiments. We observe that in the zero-shot settings, pure feature distillation is both the most efficient choice for pre-training and the most robust loss for fine-tuning.

\section{Experiments}
\label{sec:experiments}
In this section, we present the results of the models trained based on our framework and thereby explain how feature distillation enables zero-shot distillation with synthetic data. After a description of the setup we will compare our models to existing baselines which we outperform consistently. In the ablation studies, we uncover that using contrastive losses leads to students that exploit spurious features in the data, generalize poorly between synthetic and real data alongside being less robust to common corruptions of input images. Overall, this indicates that contrastive losses are a potential cause why efficient zero-shot distillation from synthetic has been an unresolved challenge.
\subsection{Setup} 
\noindent \textbf{Datasets and Hyperparameters} As introduced in Section \ref{sec:step1}, we perform feature-based pre-training on a large-scale dataset consisting of natural images for various domains. For this purpose, we select DataComp medium \cite{datacomp} and train for a single epoch. Originally, the dataset consists of 123 million images but at the time we conducted out experiments only 86$\%$ of the image URLs were still active. For fine-tuning, we target the Oxford Pets \cite{pets}, Oxford Flowers \cite{flowers}, Food-101 \cite{food} and Stanford Cars \cite{cars} to evaluate or models domain-specific datasets. These datasets are only used for testing while the actual training datasets used for training are synthetically generated based on the classes. Using the diversification strategy discussed in Section \ref{sec:diverse_synthetic}, we select a different set of five contextual dimensions and corresponding weights in the prompts to the diffusion model. More details on this selection are given in the supplementary material. For the pets, flowers and cars dataset we generate 15 options per contextual dimension while for the food dataset we use 30 as the target dataset is larger as well. This results in 265 and 1011 images per class, respectively. As the selection of options for the contextual dimensions and superclasses are relatively simple, we can use a smaller language model Llama-2 7B fine-tuned for chats \cite{llama2} and still obtain sufficiently diverse prompts. For the generation of the images, we utilize a LCM LoRA \cite{lcmlora} of Stable Diffusion XL \cite{sdxl} with a guidance scale of 0.5 and prompt weighting. As in the original CLIP paper \cite{clip} random square crops of the resized images are the only data augmentation used during training. \\For both pre-training and fine-tuning we use the same hyperparameters. We train using a batch size of 256 and a constant learning rate of $5\times 10^{-4}$ for the AdamW optimizer \cite{adamW}. All other hyperparameters were kept consistent with the CLIP training methodology \cite{clip}. One epoch of pre-training corresponds to $4.3\times10^5$ optimization steps. For fine-tuning, we perform 96 optimization epochs for all models. Even on the largest synthetic dataset this equals less than $9\%$ of the update steps done during pre-training.

\noindent \textbf{Student and Teacher Architectures} As teacher model, we employ a ViT-B/32 \cite{vit} CLIP vision encoder that has been trained on DataComp-XL, a dataset consisting of 12.8 billion image-text pairs from Common Crawl \cite{datacomp}. The corresponding text encoder follows the same architecture as described in the original CLIP paper, with 63 million parameters \cite{clip} and an embedding dimension of 512. For our student models, we utilize two different types of architectures: EfficientNets \cite{efficient_net}, which are based on convolutional neural networks, and TinyViTs \cite{efficient_net}, which are hybrid models combining convolutions and transformers. For our final results, we respectively select three models in the 5, 10, and 20 million parameter range from each architecture type. We only report intermediate results on the TinyViT with 11 million parameters. All models are systematically scaled down to improve inference speed and reduce the number of parameters while still maintaining a high capacity for representation learning. To align the output of the vision encoder with the embedding dimension of the teacher model, we apply a single linear projection head. This ensures consistent dimensionality of the embedding space between the teacher and student models.

\subsection{Zero-Shot Performance and Comparison to Baselines} 
In this section, we report the zero-shot classification accuracy of our models based on the TinyViT-11M architecture and compare them to existing benchmarks. The results are shown in Table \ref{tab:main_results}. The first reference for the performance is the teacher itself which is trained on DataComp-XL as well as the same model trained on DataComp-medium. The teacher model achieves zero-shot accuracies of over 80$\%$ on the pets, cars and food datasets as well as over 70$\%$ on the flowers dataset. Additionally, we report the performance of four TinyCLIP models that have been trained on LAION-400M \cite{laion} or YFCC-15M \cite{yfcc} datasets. These TinyCLIP models have undergone extensive training on large-scale datasets for multiple epochs, which differs from our approach. Specifically, the smallest TinyCLIP model has been exposed to six times as many images as our models, while the largest models have encountered over 120 times as many samples. The TinyCLIP models exhibit a comparable size to our models in terms of the number of parameters, even when considering the frozen text encoder which is not required for zero-shot classification. The largest TinyCLIP model has 40$\%$ fewer parameters than the ViT-B/32 CLIP model and achieves its performance up to a margin of $9\%$. The smallest TinyCLIP model features the same number of trainable parameters but has a gap of over $75\%$ to  ViT-B/32 CLIP model on the cars dataset. At the time of conducting our experiments, we were unable to compare our results with MobileCLIP \cite{mobileCLIP} and CLIP-KD \cite{clipKD} as these models are not publicly available.

\noindent We report three types of models from our framework: trained from scratch, pre-trained and finetuned. For reference, we include models that were fine-tuned on the real datasets as well. It is important to note that the reported accuracies are not zero-shot for these two cases. First, we observe that training from scratch or finetuning on synthetic data using the CLIP loss results in substantially worse performance. We assess this behavior in detail in the next section. Based on this observation, we base our framework solely on feature distillation. We find that the resulting models outperform even the largest TinyCLIP models on three of the four datasets despite having 88$\%$ less trainable parameters. Moreover, they achieve comparable performance to the teacher models with a margin of 5$\%$ on three of the four datasets. Similarly, our model outperforms the largest TinyCLIP model in these cases. The larger performance gap on the Food dataset can likely be attributed to its larger test set size, which is also evident in the TinyCLIP models. When comparing to ViT-B/32 trained on DataComp-medium, which has been trained on a comparable number of images, even our purely pre-trained models demonstrate significantly superior performance. Similarly, when pre-training a student using the CLIP loss on DataComp-medium for one epoch, the resulting zero-shot accuracies are far worse compared to the feature-based loss. This validates the data-efficiency of feature distillation for pre-training \cite{L2distillation}. We report results for top-5 validation accuracy in the supplementary material. 

\begin{table} \centering
    \caption{The upper part summarizes the baseline CLIP and TinyCLIP. ``Pre-train'' denotes pre-training on a large but non domain-specific dataset. ``Fine-tuned'' contains the results where either synthetic data (zero-shot) or real data is used for fine-tuning the pre-trained model. Gray numbers indicate that performance is not zero-shot.}
    \centering
    \label{tab:main_results}
  \resizebox{\textwidth}{!}{
    \begin{NiceTabular}{cl*{9}{c}}
        &&&& Trainable& Params.&&&\\
        & Model & Loss & Training Datasets & ImgEnc & TxtEnc & $\#$Samples Seen & Pets & Flowers & Cars & Food \\
        \midrule
        \Block{3-1}{\rotate CLIP}
        & ViT-B/32 & CLIP & DataComp-XL &  86M & 63M & 12.8B & 89.7  & 72.9  & 85.4 & 82.9 \\
        & ViT-B/32 & CLIP & DataComp-medium & 86M & 63M & 128M  & 43.1 & 29.7 & 28.0 & 41.7  \\
        & RN-50 & CLIP & openai & 86M &63M & 32$\times$400M & 85.3  & 65.2  & 54.5 & 80.8 \\  \midrule
        \Block{4-1}{\rotate TinyCLIP}
        & ViT-61M/32-29M & CLIP & LAION-400M & 61M & 29M & 38$ \times$400M & 87.3  & 64.7 & 79.1 & 73.4 \\
        & ViT-40M/32-19M & CLIP & LAION-400M & 40M&19M & 38$\times$400M & 84.4  & 61.0  & 74.2 & 71.4 \\
        & ViT-8M/16-3M & CLIP & YFCC-15M & 8M & 3M & 50$\times$15M & 45.8  & 57.4  & 8.0 & 56.2 \\
        & RN-19M-19M & CLIP & LAION-400M & 19M & 19M & 12$\times$400M & 81.0  & 56.4  & 70.1 & 66.7 \\ \midrule
        \Block{3-1}{\rotate Pre-train}
        & TinyViT-11M & CLIP & DataComp-medium &  11M & - & 110M & 10.4  & 4.2 & 5.4 & 4.7 \\  
        & TinyViT-11M & $\mathcal{L}_2$ & DataComp-medium &  11M & - & 110M & 71.4 & 39.9 & 45.0 & 52.9 \\ 
        & TinyViT-11M & $\mathcal{L}_2$ & DataComp-medium &  11M & - & 5$\times$ 110M & 78.4 & 50.0 & 58.7 & 61.1 \\ \midrule
        \Block{8-1}{\rotate Fine-tuned}
        & TinyViT-11M & $\mathcal{L}_2$ & DataComp-medium &  11M & - & 110M  &  &  &  &  \\  
        &  & CLIP & + Synthetic & &  & +1M-9M & 66.7  & 39.1  &  64.2 & 28.0 \\  
        & TinyViT-11M & $\mathcal{L}_2$ & DataComp-medium &  11M & - & 110M &  &  &  &  \\
        &  &  $\mathcal{L}_2$ & + Synthetic & & & +1M-9M  &  \textbf{87.5}  & \textbf{68.3} & \textbf{81.9} & \textbf{71.9} \\  
        & TinyViT-11M* & $\mathcal{L}_2$ & DataComp-medium &  11M & - & 110M &  &  &  &  \\  
        &  &  CLIP & + \textit{Real} Train Images &  & & +1M-7M & \textcolor{gray}{88.0} & \textcolor{gray}{90.6} &  \textcolor{gray}{90.7} & \textcolor{gray}{89.1}  \\  
        & TinyViT-11M* & $\mathcal{L}_2$ & DataComp-medium &  11M & - & 110M &  &  &  &  \\
        &  &  $\mathcal{L}_2$  & + \textit{Real} Train Images &  & & +1M-7M & \textcolor{gray}{88.7} & \textcolor{gray}{68.4} & \textcolor{gray}{83.8} & \textcolor{gray}{83.0}  \\  
        \bottomrule
    \end{NiceTabular}
    }
\end{table}

\subsection{Ablations}
Based on the observation that models trained solely using feature distillation outperform those incorporating label-based losses when fine-tuning on synthetic data, we hypothesize that the utilization of labels during fine-tuning leads the model to learn spurious features, as well as domain-specific characteristics that distinguish between synthetic and natural images. Previously, it was observed that the using the CLIP loss hinders the class-incremental generalization of students distilled on real images \cite{ood_distillation}. We examine whether this observation transfers to the synthetic to real domain shift and potential spurious features. We highlight that the class-incremental setting is different from ours, as we use a fixed set of classes. To validate our hypothesis, we conducted three experiments on the pets dataset, deliberately introducing dedicated spurious features into synthetic and real images. Additionally, we evaluate students that we fine-tuned on either real or synthetic images on test sets from the respective other domain. Like that, we aim to provide further insights into the impact of label-based losses on the transferability of the models between synthetic and natural images. Subsequently, we employ larger and smaller students with two different architecture types to assess how the performance scales with the number of trainable parameters. Additionally, we report the performance of models trained from scratch.

\begin{figure}[tb!]
  \centering
  \includegraphics[width=\textwidth]{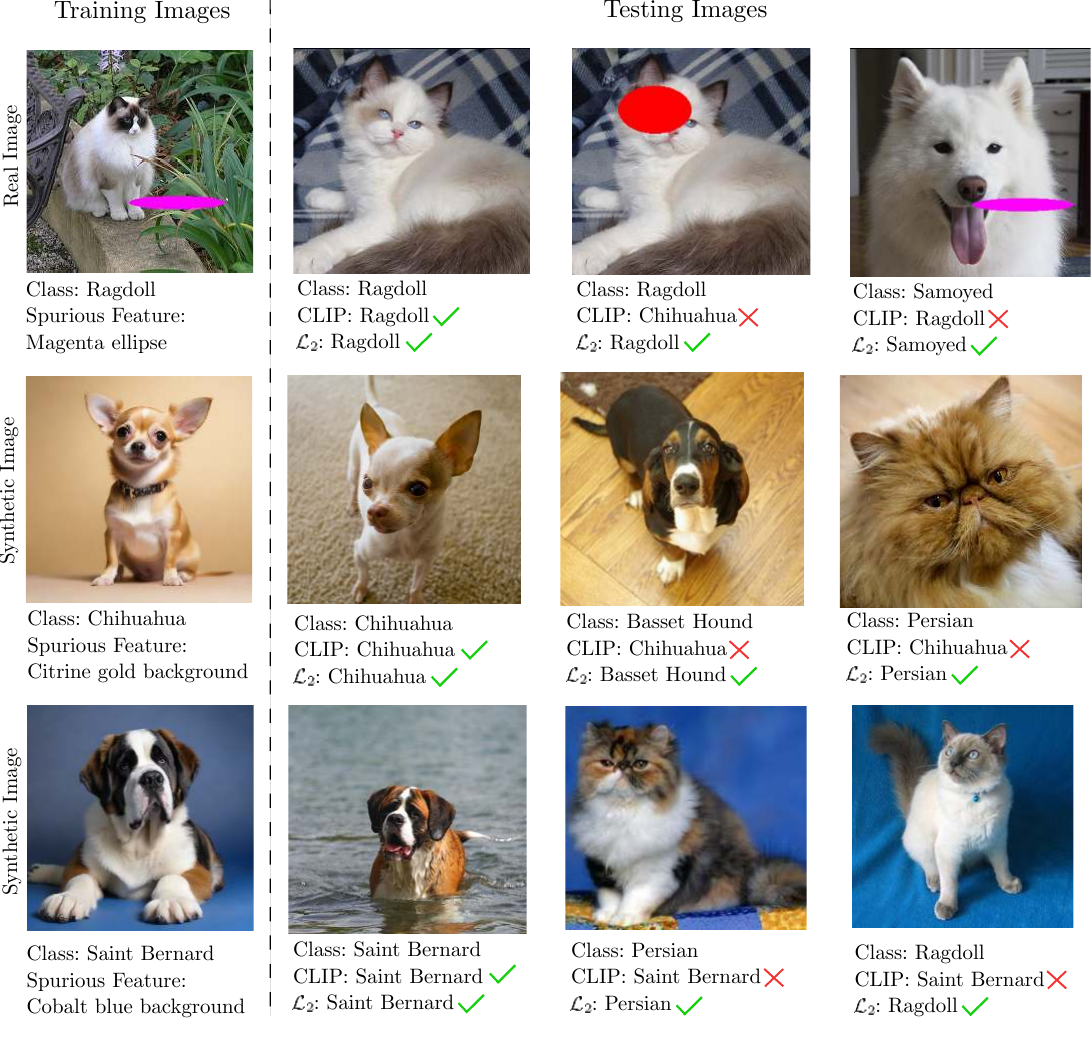}
  \caption{Examples for misclassifications of the students fine-tuned with the CLIP loss. The first row corresponds to setting with natural images. The second and third row correspond to the student trained on synthetic images. All of the test examples are classified correctly by the teacher and the student trained with $\mathcal{L}_2$ loss.}
  \label{fig:misclassification}
\end{figure}

\noindent \textbf{Natural Images with Spurious Features.} To investigate the impact of spurious features in the natural image domain, we add class-specific colored shapes to the images in the pets dataset. These shapes were added to each image in the training split, and examples can be seen in Figure \ref{fig:misclassification}. We then proceeded to fine-tune the pre-trained student models using the same hyperparameters employed during pre-training using these images. The test accuracies of the resulting models are shown in Table \ref{tab:spurious}.  We observe a slight decrease in performance on the test set without spurious features when the pre-trained students were fine-tuned with contrastive losses. This suggests that the students did not acquire any additional class-specific features during fine-tuning. However, when evaluating these fine-tuned students on a test set of real images where the colored shapes were mixed between classes, we observe a significant degradation in performance. In contrast, the students trained with the $\mathcal{L}_2$ loss achieves accuracies comparable to the dataset without spurious features on both test sets. These findings highlight the robustness of the feature loss in mitigating the negative impact of spurious features in the natural image domain.

\noindent \textbf{Synthetic Images with Spurious Features.} To investigate whether the observed behavior on real images could be replicated using synthetic ones, we generated a synthetic dataset incorporating dedicated spurious features. Specifically, we sample images where pets are positioned against a solid-colored background, with each class assigned a distinct color. The results shown in Table \ref{tab:spurious} indicate that the performance of students trained with contrastive losses deteriorates when confronted with the presence of these spurious features. Figure \ref{fig:misclassification} showcases instances of misclassifications. In contrast, the student trained with $\mathcal{L}_2$ loss exhibited a test accuracy of 84.4$\%$, which is only 5$\%$ lower than the accuracy of the teacher despite the domain gap between real and synthetic images, as well as the presence of spurious features.

\begin{table}[tb]
  \caption{
    Accuracy of the students trained on data with spurious features, introduced through adding colored shapes (real) or class-specific unicolor backgrounds (synthetic), on pets test set without spurious features. The spurious pets dataset used for testing features the same colored shapes but coupled with different classes (denoted by shuffled). Red indicates strong overfitting to trainsets with spurious features.
  }
  \label{tab:spurious}
  \centering
  \resizebox{\textwidth}{!}{
  \begin{NiceTabular}{cc*{7}{c}}
    \toprule
    Trainset & Pets Testset & \hspace*{2.0mm} $\mathcal{L}_2$ \hspace*{2.0mm} & \hspace*{1.0mm} CLIP \hspace*{1.5mm} & \hspace*{1.5mm} MP \hspace*{1.0mm} & $\mathcal{L}_2$+CLIP & $\mathcal{L}_2$+MP & Teacher  \\
    \midrule
    \Block{3-1}{\rotate Real}
    & Real Testset  & 88.9 & 60.0 & 77.3 & 88.0 & \textbf{90.0} &  89.8 \\
    & Real Testset + Shuffled Spurious Features & 88.0 & 48.7 & 51.7 & 88.0 & \textbf{88.6} & 88.7 \\
    & Real Trainset + Spurious Features & 90.3 & \textcolor{red}{96.5} & \textcolor{red}{96.5} & \textcolor{red}{95.7} & 88.3 & 89.6  \\
    \midrule
    \Block{3-1}{\rotate Synth.}
    & Real Testset & \textbf{84.4} & 24.3 & 31.0 & 81.2 & 35.6 & 89.7 \\
    & Synthetic Testset Without Spurious Features & 90.9 & 53.6 & 61.7 & \textbf{91.4} & 66.7 & 93.9  \\
    & Synthetic Trainset + Spurious Features & 94.2 & \textcolor{red}{100.0} & \textcolor{red}{100.0} & \textcolor{red}{99.3} & \textcolor{red}{100.0} & 93.3 \\
  \bottomrule
  \end{NiceTabular} }
\end{table}

\noindent \textbf{Generalization Between Synthetic and Real Images.} In order to evaluate the ability of our models to generalize between synthetic and real images, we examine the performance of models that were fine-tuned on real training datasets when tested on independently generated synthetic datasets using the same methodology as the synthetic training sets. The results are presented in Table \ref{tab:synthetic_test_pets}. The models fine-tuned with contrastive losses exhibited lower validation accuracy compared to the students trained with feature distillation. In contrast, for the models fine-tuned on synthetic data the reverse is true. Using the CLIP loss results in higher accuracy on the synthetic testset in comparison to feature distillation. This discrepancy suggests that the former models primarily learned domain-specific features of natural or synthetic images, thereby limiting their generalization capabilities between the two types.

\begin{table}[tb]
  \caption{
    Classification accuracy of the fine-tuned models for pets on a synthetic test set which was independently but identically sampled as the synthetic train set.
  }
  \label{tab:synthetic_test_pets}
  \centering
  \resizebox{\textwidth}{!}{
  \begin{tabular}{@{}lcccccccc@{}}
    \toprule
    Fine-Tuning Data & \hspace*{1.5mm} Test Data  \hspace*{1.5mm} & Teacher & Pre-Trained \hspace*{1.5mm} & \hspace*{2.5mm} $\mathcal{L}_2$ \hspace*{3.5mm} & \hspace*{2.5mm} CLIP \hspace*{1.5mm} & MP \hspace*{0.5mm} &  $\mathcal{L}_2$+CLIP \hspace*{1.5mm} & $\mathcal{L}_2$+MP  \\
    \midrule
    Real & Synthetic & 93.8 & 79.9 & \textbf{91.7} & 85.6 & 82.9 & 89.8 & 86.9 \\
    Synthetic  & Synthetic & - & - & 94.5 &\textbf{97.9} & \textbf{97.7} & 96.7 & \textbf{97.8} \\
  \bottomrule
  \end{tabular}}
\end{table}

\noindent \textbf{Robustness to Common Corruptions.} 
In order to evaluate the robustness of the models' learned representations against image perturbations, we conducted a comprehensive benchmark study. We assessed the performance of the classifiers on 15 common corruptions \cite{corruptions} at a fixed severity level of three, focusing on the pets dataset. The corresponding results are presented in Table \ref{tab:common_corruptions}. Our observations revealed that the students trained using the $\mathcal{L}_2$ feature loss demonstrate higher robustness to corruptions, regardless of whether they were trained on real or synthetic data. The distinction to the models fine-tuned with contrastive losses is particularly prominent  when training on synthetic data, where the models fine-tuned with the CLIP loss even perform worse than the purely pre-trained model. These findings provide further evidence that contrastive losses lead to learning spurious and datatype-specific features, making the models less robust to disturbances caused by common corruptions. When evaluating on synthetic test data with common corruptions, we observe that the performance drops are lower and the characteristics of the synthetic data are less disturbed by corruptions.

\begin{table}[tb]
  \caption{
    Average performance of the models, which were fine-tuned on real data, on the pets testset under 15 common corruptions with severity 3.
  }
  \label{tab:common_corruptions}
  \centering
  \resizebox{\textwidth}{!}{
  \begin{tabular}{@{}lcccccccc@{}}
    \toprule
    Fine-Tuning Data & \hspace*{1.5mm} Test Data \hspace*{1.5mm} & Teacher & Pre-Trained \hspace*{1.5mm} & \hspace*{2.5mm} $\mathcal{L}_2$ \hspace*{3.5mm} & \hspace*{2.5mm} CLIP \hspace*{1.5mm} & MP \hspace*{0.5mm} &  $\mathcal{L}_2$+CLIP \hspace*{1.5mm} & $\mathcal{L}_2$+MP  \\
    \midrule
    Real & Real & 78.2 & 52.4 & 73.6 & 65.3 & 64.8 & \textbf{77.7} & 66.7 \\
    Synthetic  & Real & - & - & \textbf{66.2} & 40.0 & 37.3 & 63.3 & 38.2 \\
    Real & Synthetic & 93.6 & 73.5 & 93.8 & 94.5 & 94.2 & \textbf{95.2} & 93.2 \\
    Synthetic & Synthetic & - & - & 90.8 & 83.5 & 83.4 & \textbf{91.2} & 85.4 \\
  \bottomrule
  \end{tabular}}
\end{table}

\begin{figure}[tb]
  \includegraphics[width=\textwidth]{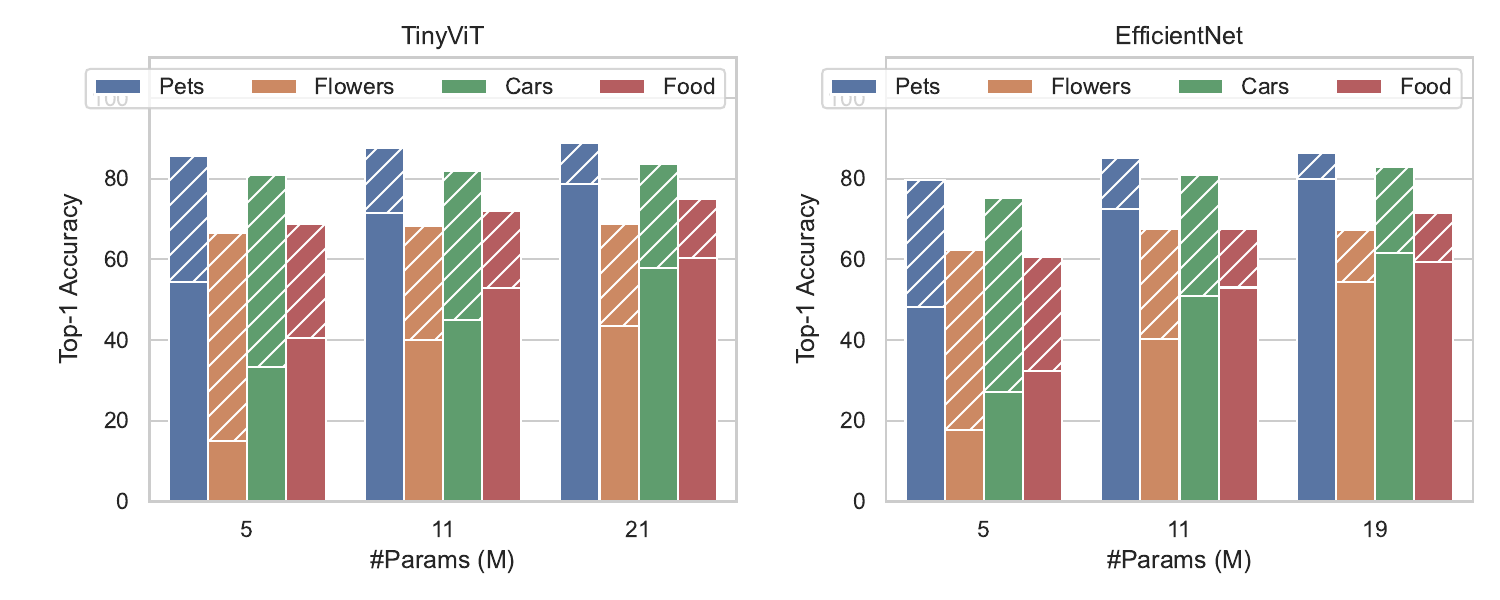}
  \caption{
    Zero-shot classification performance of the students after pre-training on DataComp medium for one epoch and after finetuning on the synthetic datasets for 96 epochs (hatched). All experiments were performed only using feature distillation.
  }
  \label{fig:other_students}
\end{figure}

\noindent \textbf{Student Model Size.} 
Alongside the results of a TinyViT-11M in Table \ref{tab:main_results}, we report the zero-shot performance of five additional students after pre-training and fine-tuning using feature distillation in Figure \ref{fig:other_students}. As expected, there is a general trend of improved performance with increasing model size. Yet, we observe that the effectiveness of zero-shot fine-tuning is more pronounced for smaller models compared to larger ones: after fine-tuning, the difference in performance between the largest and smallest models is around 10$\%$ to 15$\%$, while after purely pre-training it is as high as 30$\%$. These findings highlight that zero-shot distillation is  particularly effective for smaller models since it allows to adapt them to the target domain, without requiring real in-domain data.

\begin{figure}[tb]
  \includegraphics[width=\textwidth]{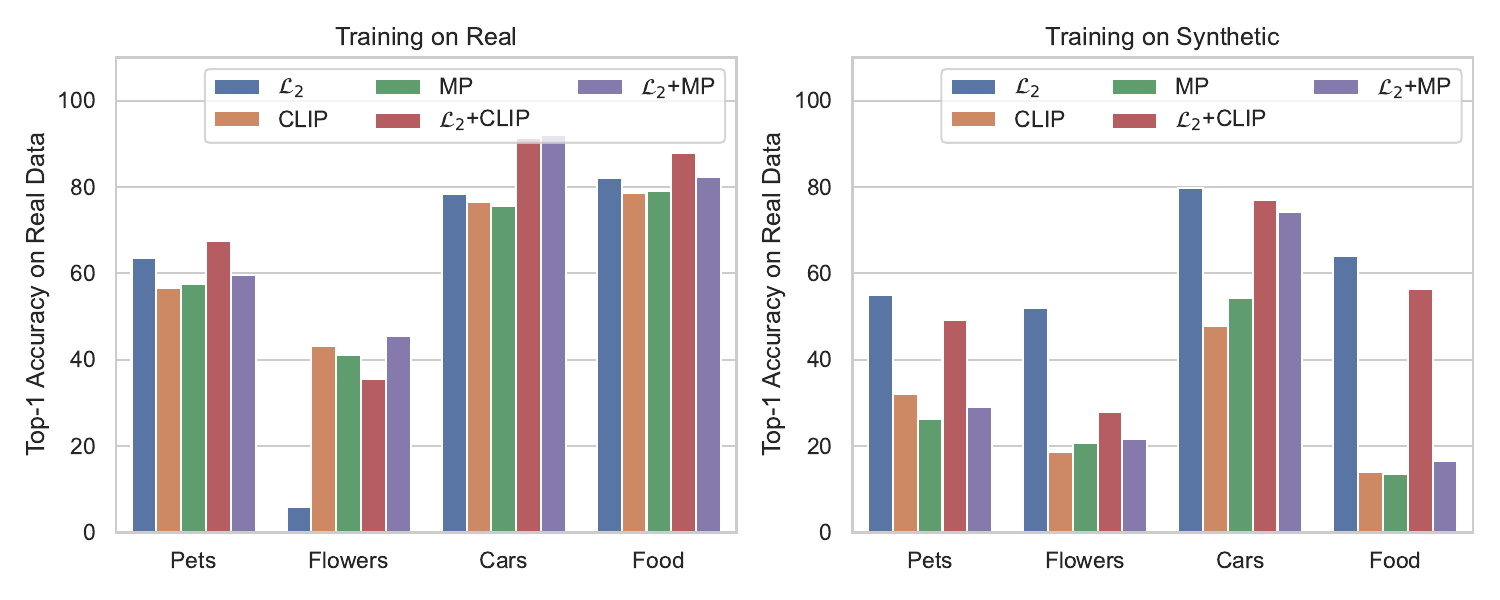}
  \caption{
    Classification performance of the models after training from scratch for 96 epochs on either real or synthetic images.
  }
  \label{fig:scratch}
\end{figure}

\noindent \textbf{Training From Scratch.} We investigate the accuracy of models trained from scratch, both on real and synthetic images in \cref{fig:scratch}. On real data, using a sum of the $\mathcal{L}_2$ and a contrastive loss indeed results in the best performance as observed in CLIP-KD \cite{clipKD}. The low accuracy of the feature distilled student on the flowers dataset can be attributed to the small training set with only 10 images per class. On synthetic training data, the $\mathcal{L}_2$ loss consistently performs best while the CLIP loss or the MP loss results are substantially worse. Combing the $\mathcal{L}_2$ with one of the contrastive losses does not yield any performance gains in contrast to purely feature-based distillation. This is in line with the behavior observed during fine-tuning in \cref{tab:main_results}.

\section{Conclusion}
In this work, we introduced a framework for zero-shot distillation of small CLIP image encoders based on synthetic images. We make the surprising observation that contrastive losses are a potentially detrimental factor for generalization capabilities of models between synthetic and real data, due to the exploitation of spurious features. By employing a pure image feature-based distillation loss, we successfully mitigate this limitation. As a result, we are able to train models that surpass the current state-of-the-art for zero-shot CLIP distillation, while featuring fewer parameters and  not using labeled target domain images.

\noindent \textbf{Limitations and Future Work.} The presented results are based on a ViT-B/32 teacher; future work could explore the benefit of larger teachers for constant-sized students. Furthermore, the potential of our small-scale image encoders beyond zero-shot settings could be explored, for instance in architectures that use CLIP image encoders such as BLIP-2 \cite{blip2} or LLava \cite{llava1,llava2,llava3}. Moreover, the current framework is limited to classification tasks. To broaden its applicability, future research could extend the framework to encompass other computer vision tasks such as object detection or image segmentation. 
\newpage
\noindent \textbf{Acknowledgements.} We would like to thank Nicole Finnie for helpful discussion on pre-training CLIP models. We also thank the European Laboratory for Learning and Intelligent Systems (ELLIS) for supporting
Niclas Popp.

%
%
\bibliographystyle{splncs04}
\bibliography{egbib}

\newpage

\appendix
\newpage
\section{Supplementary Material}
\subsection{Details of the Synthetic Data Generation}
\begin{figure}
\centering
  \includegraphics[width=0.9\textwidth]{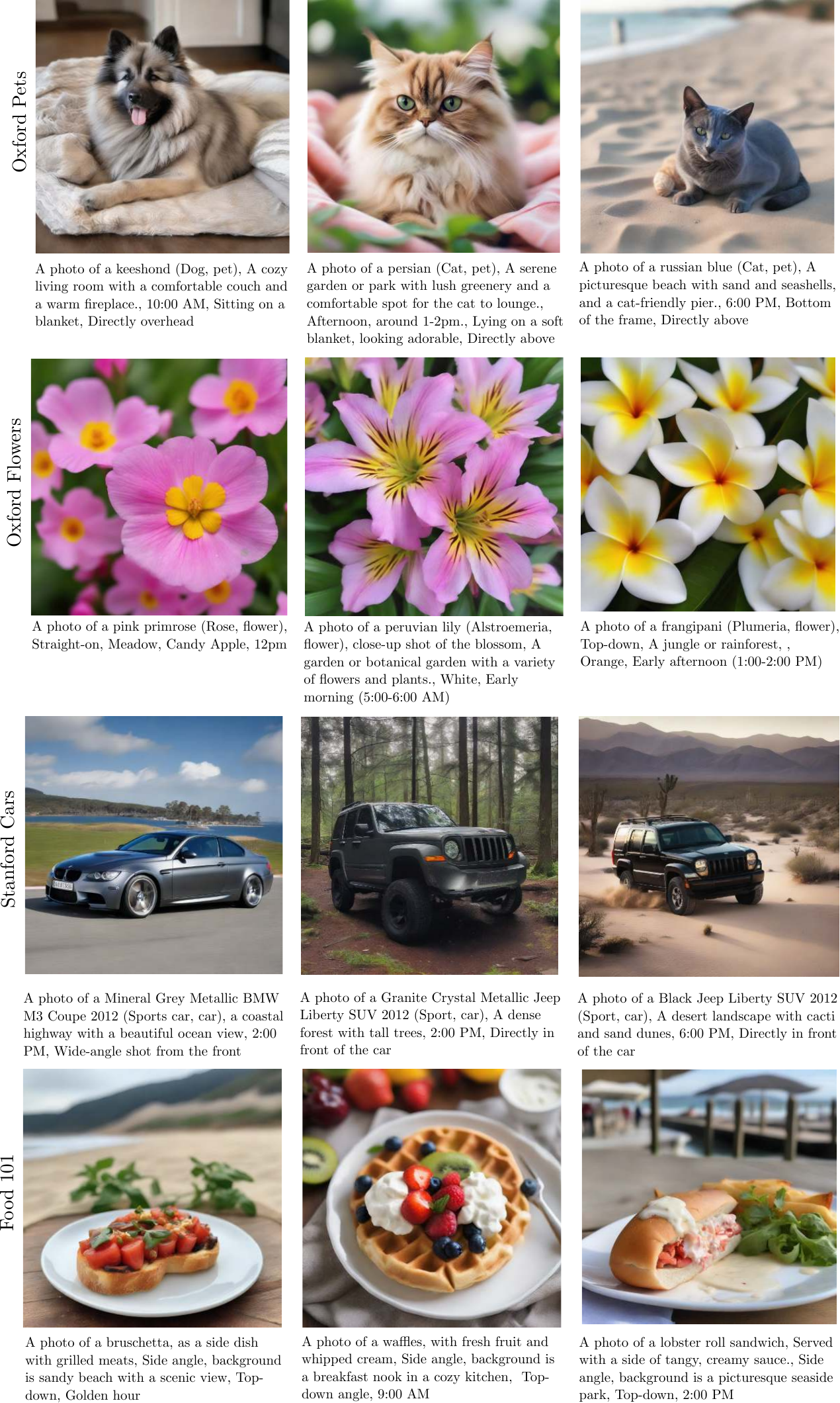}
  \caption{
    Examples for synthetic images from all four datasets together with the prompts used to generate them.
  }
  \label{fig:examples}
\end{figure}

\begin{figure}[tb]
\centering
  \includegraphics[width=\textwidth]{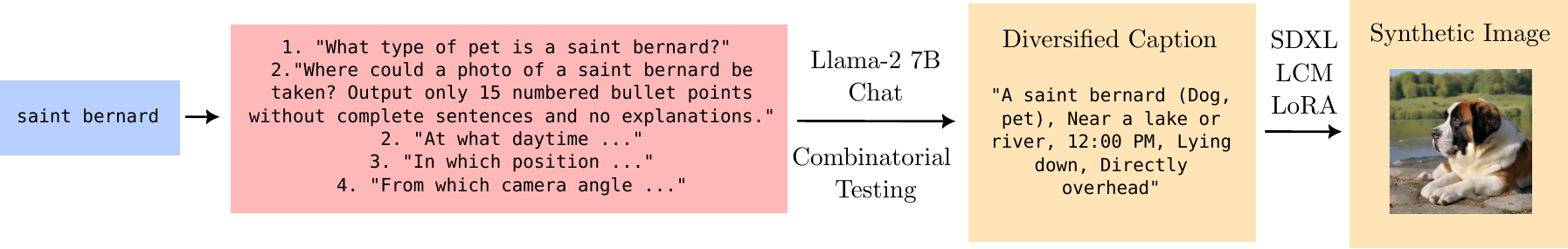}
  \caption{
    The figure illustrates the process for generating synthetic images for the example class "saint bernard" from the pets dataset.
  }
  \label{fig:pipe_examples}
\end{figure}

In this section, we provide further details on the synthetic data generation and the diversification process. As mentioned in Section \ref{sec:diverse_synthetic}, the prompts used to synthesize the images are based on the classnames and additional information given by an LLM. For each class, we ask the language model to provide information with respect to four contextual dimensions as well as a superclass. The contextual dimensions are dataset specific and are summarized in Table \ref{tab:contextual_dimensions}. Figure \ref{fig:pipe_examples} shows a concrete example for a class from the pets dataset. For each of the contextual dimensions we collect 15 or 30 options from Llama 2 7B fine-tuned for chats \cite{llama2}. The larger number of options for the food dataset is used to accommodate its larger test set. In Table \ref{tab:datasets} we summarize the sizes of the real target datasets. Instead of using all possible combinations of options for the contextual dimensions to generate the prompts, we use combinatorial testing \cite{comb_test1,comb_test2}. This approach is inspired by a recent work on systematic error identification \cite{promptattack}. It reduces the number of images per class while ensuring that the prompts systematically cover the diversity contained in the answers from the LLM. For example, in case of 15 options per contextual dimension, this results in 265 images per class instead of 50625. The prompts are a comma separated list of the selected options, which are weighted to accommodate for contextual dimensions that are more or less important for certain domains. These weighted prompts are then used as input for a diffusion model. Specifically, we use Stable Diffusion XL \cite{sdxl} LCM LoRA \cite{lcmlora}. To ensure sufficient image quality and diversity we employ a guidance scale of 0.5 and 6 inference steps. Further example images can be found in Figure \ref{fig:examples}. These also showcase some of the known problems with diffusion models such as parts of the prompts which are missing in the image \cite{diffusion_problems} as in the first example for the food dataset.

\begin{table}[h!]
  \caption{
    Contextual dimensions and prompt weights for the diversified data generation
  }
  \label{tab:contextual_dimensions}
  \centering
  \resizebox{\textwidth}{!}{
  \begin{tabular}{@{}lcccc@{}}
    \toprule
     & Weight of & Contextual Dimensions  & Options & Images \\ 
    Dataset & Classname & and Weights in Braket & per Dimension & per Class\\ 
    \midrule
    Pets  & 1.5 & superclass (1.2), locations,  & 15 & 265 \\
    & & position, daytime, camera angle & &\\
    Flowers & 1.2 & superclass, color, locations,  & 15 & 265\\
    & & daytime (0.1), camera angle (0.1) & &\\
    Cars & 1.0 & superclass, locations,  & 15 & 265\\
    & & color, daytime, camera angle & &\\
    Food & 1.2 & superclass, locations, way of serving (1.5),  & 30 & 1011 \\
    & &  daytime(0.1), camera angle(0.1) & &\\
  \bottomrule
  \end{tabular}}
\end{table}
\newpage

\begin{table}[tb]
  \caption{
    Overview over the size of the real target datasets
  }
  \label{tab:datasets}
  \centering
  \begin{tabular}{@{}lccc@{}}
    \toprule
    Dataset & $\#$classes & $\#$training images & $\#$test images  \\
    \midrule
    Pets & 37 & 3680 & 3669 \\
    Flowers & 102 & 1020 & 6149 \\
    Cars & 196 & 8144 & 8041 \\
    Food & 101 & 75750 & 25250 \\
  \bottomrule
  \end{tabular}
\end{table}

\subsection{Overview over (Zero-Shot) Baselines}
To complement the framework presented in Section \ref{sec:experiments}, we situate existing baselines with respect to the discussed components in Table \ref{tab:baseline_overview}. We state whether pre-training and/or fine-tuning is used on synthetic or natural images or both. Data diversification describes if the approach uses prompt or images diversification for the synthetic data. We note that our framework unifies existing pipelines.

\begin{table}[tb]
  \caption{
    Comparison of existing training approaches to our zero-shot distillation framework. Similar to our framework, the DM-KD pipeline only relies on synthetic images for training on specific target datasets. However, the teacher was trained on the real images, which is not possible in a zero-shot setting. This is symbolized by $^*$.
  }
  \label{tab:baseline_overview}
  \centering
  \resizebox{\textwidth}{!}{
  \begin{tabular}{@{}lcccccccc@{}}
    \toprule
     & Evaluation & Pre-  & Fine- & Natural  & Synthetic & Data & \\ 
    Name & Metric & Training  & Tuning &  Images & Images & Diversification & Loss \\ 
    \midrule
    StableRep \cite{stablerep} & Linear probe, few-shot & \checkmark & & &\checkmark & & MP\\
    SynCLR \cite{modelsvsdata} & Linear probe & \checkmark & & &\checkmark &\checkmark & MP \\
    SynthCLIP \cite{synthCLIP} & Linear probe, few-shot & \checkmark & & & \checkmark & \checkmark & CLIP\\
    Fake it till you make it \cite{fake} & Zero-Shot Acc. & \checkmark & & & \checkmark &  & Cross-Entropy \\
    Diversify don't finetune \cite{diversify} & Accuracy  & \checkmark & & \checkmark & \checkmark & \checkmark & Custom\\
    \cite{syntheticImageNet} & Accuracy  & \checkmark & & \checkmark & \checkmark & & Cross-Entropy \\
    DM-KD \cite{syntheticDistillation} & Accuracy$^*$ & & \checkmark & & \checkmark & & KD \cite{ogdistillation} \\
    TinyCLIP \cite{tinyCLIP} & Zero-Shot Acc. & \checkmark & & \checkmark & & & CLIP, Affinity Mapping \\
    MobileCLIP \cite{mobileCLIP} & Zero-Shot Acc. & \checkmark & & \checkmark & & & CLIP, Affinity Mapping \\
    \textbf{Zero-Shot Distillation} (Ours) & Zero-Shot Acc. & \checkmark & \checkmark & \checkmark & \checkmark & \checkmark & $\mathcal{L}_2^{feature}$  \\
  \bottomrule
  \end{tabular}}
\end{table}

\subsection{Top-5 Test Accuracies}
\begin{figure}[tb]
  \includegraphics[width=\textwidth]{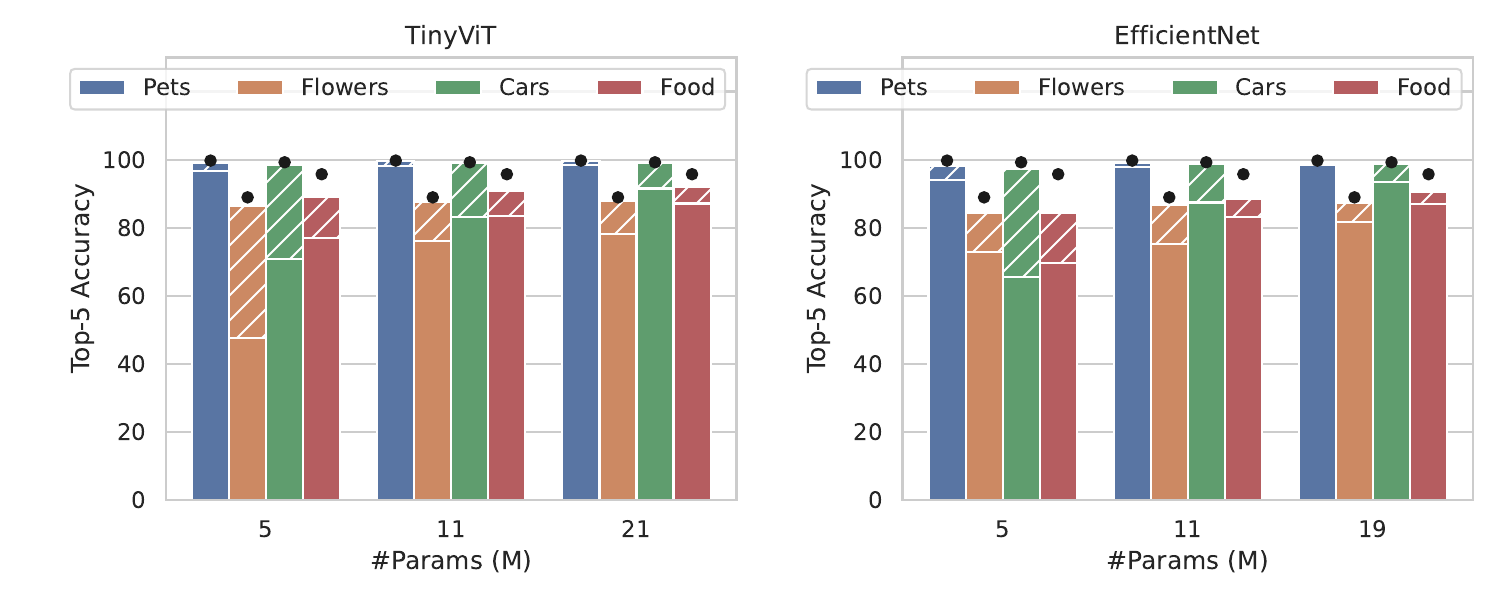}
  \caption{
    Top-5 accuracy of the models  after pre-training on DataComp medium for one epoch and after fine-tuning on the synthetic datasets for 96 epochs (hatched). The black dots indicate the top-5 accuracy of the teacher.
  }
  \label{fig:top5}
\end{figure}

To complement the results in Section \ref{sec:experiments}, we provide the top-5 accuracies of the pre-trained and fine-tuned models. Figure \ref{fig:top5} visualizes the results. The trends mirror the observations from the Top-1 accuracy with an even smaller gap to the teacher.

\subsection{Training Accuracies}
In addition to the test accuracies reported in the main paper, we report the training accuracies of the fine-tuned models in Table \ref{tab:training_accuracy}. We observe that the models trained with a contrastive loss achieve higher training accuracy than the feature distilled students. This holds in particular on synthetic data. In combination with the results from Section \ref{sec:experiments}, this underlines our hypothesis that contrastive losses lead to learning domain specific or spurious features over actual class or object specific features.

\begin{table}[tb]
  \caption{
    Accuracy of the fine-tuned models on the datasets they were trained on.
  }
  \label{tab:training_accuracy}
  \centering
  \begin{NiceTabular}{@{}lccccccccc@{}}
    \toprule
     &  & Pets & Pets & Flowers & Flowers & Cars & Cars & Food & Food \\
    Trainset & Loss & Top-1 &  Top-5 &  Top-1 & Top-5 & Top-1 & Top-5 & Top-1 & Top-5\\ \midrule
    \Block{5-1}{\rotate Real} 
    & $\mathcal{L}_2$ & 87.7 & 99.0 & 70.3 & 89.3 & 83.8 & 99.1 & 79.3 & 93.7 \\
    & CLIP & 89.7 & 98.5 & 97.3 & 100.0 & 90.6 & 100.0 & 97.6 & 99.2  \\
    & MP & 89.1 & 98.5 & 97.5 & 100.0 & 90.7 & 100.0 & 97.5 & 99.2  \\
    & $\mathcal{L}_2$+CLIP & 91.2 & 99.9 & 90.6 & 98.0 & 92.2 & 100.0 & 90.8 & 98.3  \\
    & $\mathcal{L}_2$+MP & 100.0 & 100.0 & 97.7 & 100.0 & 92.8 & 100.0 & 98.5 & 99.6 \\ 
    \midrule
    \Block{5-1}{\rotate Synthetic} 
    & $\mathcal{L}_2$  & 95.3 & 100.0 & 68.0 & 90.1 & 75.9 & 95.7 & 84.5 & 96.3 \\
    & CLIP & 100.0 & 100.0 & 97.8 & 98.9 & 90.0 & 99.2 & 99.7 & 100.0 \\
    & MP  & 99.9 & 100.0 & 99.5 & 100.00 & 87.6 & 98.4 & 99.7 & 100.0 \\
    & $\mathcal{L}_2$+CLIP & 97.9 & 100.0 & 85.8 & 96.9 & 91.0 & 99.7 & 93.2 & 98.8 \\
    & $\mathcal{L}_2$+MP & 100.0 & 100.0 & 99.6 & 100.0 & 94.7 & 100.0 & 99.9 & 100.0 \\
  \bottomrule
  \end{NiceTabular}
\end{table}

\subsection{Contrastive Image Loss}
As an alternative  to the feature-based $\mathcal{L}_2$ loss, we test a contrastive loss that is purely based on the image feature of the student and teacher. Using the notation from Section \ref{sec:loss}, it is defined as
\begin{equation}\label{eq:contrastive_single_modal}
    \mathcal{L}_{\textrm{contrastive}}^{\textrm{image}} = \sum_{i=1}^{N}-\textrm{log}\frac{\textrm{exp}(\langle \textbf{I}_i^S,\textbf{I}_i^T \rangle/\tau)}{\sum_{k=1}^{N} \textrm{exp}(\langle \textbf{I}_i^S,\textbf{I}_k^T \rangle/\tau)} 
\end{equation}
where $\tau$ denotes a learnable temperature parameter. We use this loss both for pre-training and fine-tuning a TinyViT 11M with the same setup as for the $\mathcal{L}_2$ loss in Section \ref{sec:experiments}. The results are reported in Table \ref{tab:I2I}. We observe that for pre-training, the contrastive image loss results in better performance in comparison to the $\mathcal{L}_2$ loss, while for fine-tuning it is the other way around. Yet, the contrastive image loss still clearly outperforms the CLIP loss when fine-tuning on synthetic data. This validates our observation from Section \ref{sec:experiments} that using a purely feature based loss improves the generalization between synthetic and real data.
\begin{table}[bp!]
  \caption{
    Accuracy of the models that were pre-trained and fine-tuned using distillation with the contrastive image loss. The differences to $\mathcal{L}_2$ feature distillation and training using the CLIP loss are shown in gray.
  }
  \label{tab:I2I}
  \centering
  \resizebox{\textwidth}{!}{
  \begin{NiceTabular}{cc*{5}{c}}
    \toprule
    Trainset & Training & Loss & \hspace*{2.0mm} Pets\hspace*{2.0mm} & \hspace*{1.0mm} Flowers \hspace*{1.5mm} & \hspace*{1.5mm} Cars \hspace*{1.0mm} & Food  \\
    \midrule
    \Block{8-1}{\rotate Real}
    & Pre-Trained & Contrastive Image  & 72.8 & 39.6 & 46.5 & 54.5   \\
    & & \textcolor{gray}{Difference to $\mathcal{L}_2$} & \textcolor{gray}{+1.4} & \textcolor{gray}{+0.5} & \textcolor{gray}{+1.5} & \textcolor{gray}{+1.4} \\
    & & \textcolor{gray}{Difference to Contrastive} & \textcolor{gray}{+52.4} & \textcolor{gray}{+35.4} & \textcolor{gray}{+41.1} & \textcolor{gray}{+49.8} \\
     & & \textcolor{gray}{Image-Text (CLIP)} &  &  &  & \\
    & Fine-Tuned & Contrastive Image & 83.9 & 64.9 & 81.4 & 80.4 \\
    & & \textcolor{gray}{Difference to $\mathcal{L}_2$} & \textcolor{gray}{-4.8} & \textcolor{gray}{-3.5} & \textcolor{gray}{-2.4} & \textcolor{gray}{-2.6} \\
    & & \textcolor{gray}{Difference to Contrastive} & \textcolor{gray}{-5.5} & \textcolor{gray}{-25.7} & \textcolor{gray}{-9.3} & \textcolor{gray}{-2.6} \\
     & & \textcolor{gray}{Image-Text (CLIP)} &  &  &  & \\
    \midrule
    \Block{4-1}{\rotate Synth.}
    & Fine-Tuned & Contrastive Image & 80.5 & 57.7 & 79.3 & 68.8 \\
    & & \textcolor{gray}{Difference to $\mathcal{L}_2$} & \textcolor{gray}{-7.0} & \textcolor{gray}{-10.6} & \textcolor{gray}{-2.6} & \textcolor{gray}{-3.1} \\
    & & \textcolor{gray}{Difference to Contrastive} & \textcolor{gray}{+13.8} & \textcolor{gray}{+18.6} & \textcolor{gray}{+15.1} & \textcolor{gray}{+40.8} \\ & & \textcolor{gray}{Image-Text (CLIP)} &  &  &  & \\
  \bottomrule
  \end{NiceTabular} }
\end{table}
\subsection{ImageNet-100 Fine-Tuning}
As an addition to the domain specific datasets discussed in the main paper, we fine-tune the models on ImageNet-100 \cite{ImageNet100} which is a subset of ImageNet-1k consisting of 100 classes with various objects that do not necessarily belong to a similar domain. To generate synthetic images, we use the same setup as for the datasets in Section \ref{sec:experiments} with prompt diversification and 30 options per contextual dimension. This equals 1011 images per class. The contextual dimensions are the same as for the cars dataset but without any prompt weighting. Instead of using the simple zero-shot prompts "a photo of a ..., which is a type of ...", we use the prompt ensembles consisting of 79 templates proposed by Radford et al. \cite{clip}. These prompt templates are also used for the diversified prompts by starting with a randomly chosen one before listing the classname and the options for the contextual dimensions. The results are shown in Table \ref{tab:IN}. For the contrastive losses, we observe the same trends as for the domain specific datasets. When fine-tuning on real data, the resulting test performance exceeds the teacher while fine-tuning with synthetic data deteriorates the accuracy in contrast to pure pre-training. For the $\mathcal{L}_2$ loss this is not the case. Yet, fine-tuning with feature distillation yields almost the same zero-shot accuracy as pure pre-training. This is different to the domain specific datasets where we could observe a consistent improvement. This is likely due to the larger diversity in the real test images which is not sufficiently captured by the synthetic training data.
\begin{table}[tb]
  \caption{
    Zero-Shot accuracy of a TinyViT 11M pre-trained for one epoch on DataComp medium and fine-tuned on ImageNet-100 using either real or synthetic data. Using the CLIP loss with synthetic data decreases the performance in contrast to pure pre-training. Fine-tuning through $\mathcal{L}_2$ feature distillation yields a slight improvement.
  }
  \label{tab:IN}
  \centering
  \resizebox{\textwidth}{!}{
  \begin{tabular}{@{}lccccccc@{}}
    \toprule
    Training Data & Teacher & Pre-Trained \hspace*{1.5mm} & \hspace*{2.5mm} $\mathcal{L}_2$ \hspace*{3.5mm} & \hspace*{2.5mm} CLIP \hspace*{1.5mm} & MP \hspace*{0.5mm} &  $\mathcal{L}_2$+CLIP \hspace*{1.5mm} & $\mathcal{L}_2$+MP  \\
    \midrule
    Real & 86.1 & 74.3 & 81.8 & 87.0 & 87.7 & 87.2 & \textbf{89.3}  \\
    Synthetic & - & - & 74.0 & 53.2 & 52.4 & \textbf{74.2} & 68.2   \\
  \bottomrule
  \end{tabular}}
\end{table}
\subsection{Linear Probing}
To evaluate the linear probe accuracy of the teacher as well as the TinyViT 11M models pre-trained and fine-tuned on the pets dataset, we fit a linear classifier based on the unnormalized image features after the projection head. The classifier is fitted either using synthetic or real data, where only the case of synthetic data corresponds to the zero-shot setting. The hyperparameter
sweeps for the regularization are performed on a validation split as in the original CLIP paper \cite{clip}. The results are shown in Table \ref{tab:linear_probe}. For the models fine-tuned on synthetic data, the performance is 8 to 10 $\%$ worse when probing with synthetic data in comparison to fitting the linear classification head with real data. This highlights that using linear probing based on real data improves the linear classification accuracy by a substantial margin. In contrast, the true zero-shot linear classifiers, where the classification head is fitted using synthetic data, perform comparable to or worse than pure zero-shot classification using the similarity between the image and prompt embeddings. In contrast to our framework, previous works \cite{stablerep,modelsvsdata,synthCLIP} mainly focus on the linear accuracy where the classification head is fitted with real data instead of targeting the true zero-shot setting without any real data which is the more difficult task to accomplish.
\begin{table}[tb]
  \caption{
    Linear probe accuracy of the teacher as well as TinyViT 11Ms models pre-trained for one epoch on DataComp medium and fine-tuned on pets. The linear accuracy of the models that were fine-tuned with synthetic data is increased substantially by probing with real data compared to probing with synthetic data.
  }
  \label{tab:linear_probe}
  \centering
  \resizebox{\textwidth}{!}{
  \begin{tabular}{@{}lcccccccc@{}}
    \toprule
    Fine-Tuning Data & Probing Data & Teacher & Pre-Trained \hspace*{1.5mm} & \hspace*{2.5mm} $\mathcal{L}_2$ \hspace*{3.5mm} & \hspace*{2.5mm} CLIP \hspace*{1.5mm} & MP \hspace*{0.5mm} &  $\mathcal{L}_2$+CLIP \hspace*{1.5mm} & $\mathcal{L}_2$+MP  \\
    \midrule
    Real & Real & 90.4 & 81.6 & 89.8 & 88.3 & 88.9 & 92.1 & 90.2  \\
    Real & Synthetic & 84.0 & 71.8 & 82.1 & 87.7 & 88.3 & 87.7 & 88.9  \\
    Synthetic & Real & - & - & 89.6 & 73.7 & 72.6 & 90.1 & 75.3 \\
    Synthetic & Synthetic & - & - & 80.9 & 64.5 & 65.0 & 82.8 & 65.8 \\
  \bottomrule
  \end{tabular}}
\end{table}
\subsection{Performance Under Domain Shift} 
\begin{figure}[tb]
\centering
  \includegraphics[width=\textwidth]{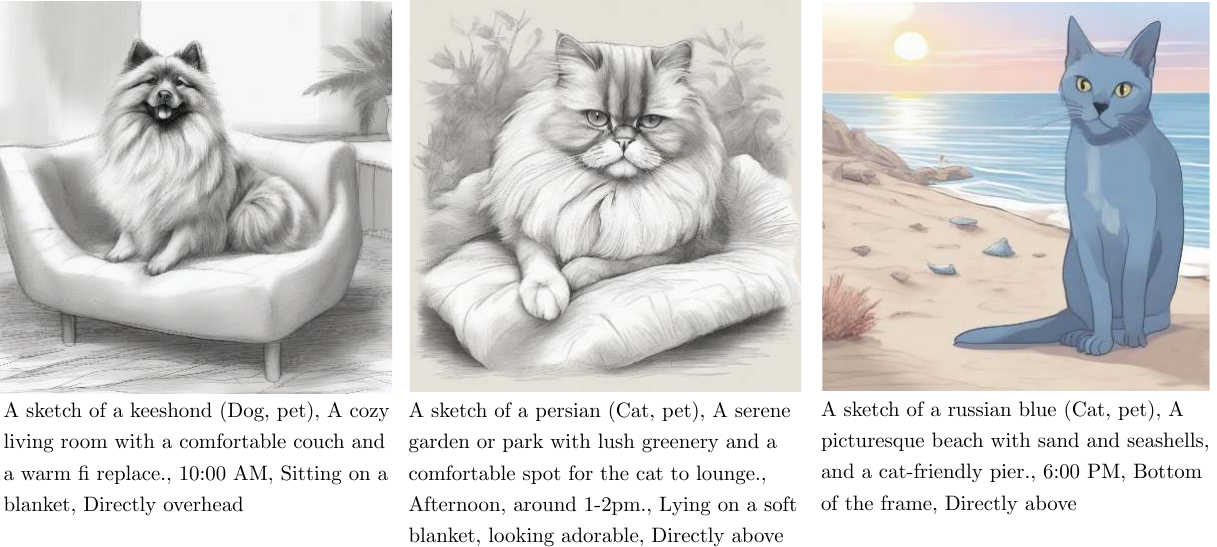}
  \caption{
    Examples from the synthetic test set which comprises of sketches.
  }
  \label{fig:skectche}
\end{figure}

In addition to the synthetic-real domain gap, we investigate the accuracy of the fine-tuned models under dedicated domain shift. This setting is comparable to the experiments performed by Zhou et al. \cite{ood_robustness_distillation}. However, our setup requires no adversarial training. We sample a dataset of images that utilizes the prompt template "a sketch of a ..., which is a type of ..." instead of "a photo of a ..., which is a type of ..." to generate sketches rather than photorealistic images. The remainders of the prompts are identical to those of the independent synthetic test set. Figure \ref{fig:skectche} illustrates example images from this dataset. For zero-shot classification, we continue to employ "a photo of a ..., which is a type of ..." to simulate unknown domain shift. The results are presented in Table \ref{tab:domainshift}. When comparing to the testsets without dedicated domain shifts, we observe that the feature distilled student achieves the lowest decrease in performance, substantially outperforming the models that were fine-tuned using purely contrastive losses. Interestingly, the drop in accuracy is larger when using synthetic in comparison to real data for fine-tuning where the test set feature the domain shift from photos to sketches but not from synthetic to real images. Additionally, we test the models fine-tuned on ImageNet-100 on the corresponding classe of ImageNet Sketch \cite{IN_sketch}. The results are shown in Table \ref{tab:domainshift_IN}. In this case, the students distilled with the feature loss are again more robust against the shift to sketches. Our observations on domain shift mirror the behavior in the class-incremental setting \cite{ood_distillation}. There, the models are trained on a subset of the real training images that contains only a fraction of the overall classes and are evaluated on the set of remaining classes. In our case, however, we use a fixed number of classes.
\begin{table}[tb]
  \caption{
    Accuracy of the teacher as well as TinyViT 11Ms models pre-trained for one epoch on DataComp medium and fine-tuned on synthically generated pets test data which comprises of sketches. The feature-distilled student exhibit a substantially smaller drop in performance compared to the contrastively fine-tuned models.
  }
  \label{tab:domainshift}
  \centering
  \resizebox{\textwidth}{!}{
  \begin{tabular}{@{}lccccccc@{}}
    \toprule
    Fine-Tuning Data & Teacher & Pre-Trained \hspace*{1.5mm} & \hspace*{2.5mm} $\mathcal{L}_2$ \hspace*{3.5mm} & \hspace*{2.5mm} CLIP \hspace*{1.5mm} & MP \hspace*{0.5mm} &  $\mathcal{L}_2$+CLIP \hspace*{1.5mm} & $\mathcal{L}_2$+MP  \\
    \midrule
    Real & 89.3 & 71.3 & 83.3 & 70.4 & 72.6 & 83.8 & 73.0 \\
    \textcolor{gray}{Difference to real test data} & \textcolor{gray}{-0.4} & \textcolor{gray}{-7.1} & \textcolor{gray}{-5.4} & \textcolor{gray}{-17.6} & \textcolor{gray}{-16.4} & \textcolor{gray}{-13.9} & \textcolor{gray}{-17.6}\\
    Synthetic & - & - & 86.9 & 76.0 & 75.2 & 87.9 & 81.9 \\
    \textcolor{gray}{Difference to synthetic test data} & & \textcolor{gray}{} & \textcolor{gray}{-7.6} & \textcolor{gray}{-21.5} & \textcolor{gray}{-22.5} & \textcolor{gray}{-8.8} & \textcolor{gray}{-15.9}\\
  \bottomrule
  \end{tabular}}
\end{table}

\begin{table}[tb]
  \caption{
    Zero-Shot accuracy of the models fine-tuned on ImageNet-100 when tested on the 100 corresponding classes in ImageNet Sketch \cite{IN_sketch}. The students distilled using feature distillation exhibit better performance as well as lower degration in comparison to the accuracy on standard ImageNet-100
  }
  \label{tab:domainshift_IN}
  \centering
  \resizebox{\textwidth}{!}{
  \begin{tabular}{@{}lccccccc@{}}
    \toprule
    Fine-Tuning Data & Teacher & Pre-Trained \hspace*{1.5mm} & \hspace*{2.5mm} $\mathcal{L}_2$ \hspace*{3.5mm} & \hspace*{2.5mm} CLIP \hspace*{1.5mm} & MP \hspace*{0.5mm} &  $\mathcal{L}_2$+CLIP \hspace*{1.5mm} & $\mathcal{L}_2$+MP  \\
    \midrule
    Real & 76.3 & 54.1 & 57.5 & 49.4 & 51.0 & \textbf{60.6} & 56.8 \\
    \textcolor{gray}{Difference to real test data} & \textcolor{gray}{-14.1} & \textcolor{gray}{-20.2} & \textcolor{gray}{-24.3} & \textcolor{gray}{-38.0} & \textcolor{gray}{-36.7} & \textcolor{gray}{-27.2} & \textcolor{gray}{-32.5}\\
    Synthetic & - & - & \textbf{56.3} & 34.0 & 25.6 & 55.7 & 49.8 \\
    \textcolor{gray}{Difference to real test data} &  &  & \textcolor{gray}{-17.7} & \textcolor{gray}{-19.2} & \textcolor{gray}{-26.8} & \textcolor{gray}{-18.5} & \textcolor{gray}{-18.4}\\
  \bottomrule
  \end{tabular}}
\end{table}

\subsection{Simple Prompts} 
\begin{figure}[tb]
\centering
  \includegraphics[width=\textwidth]{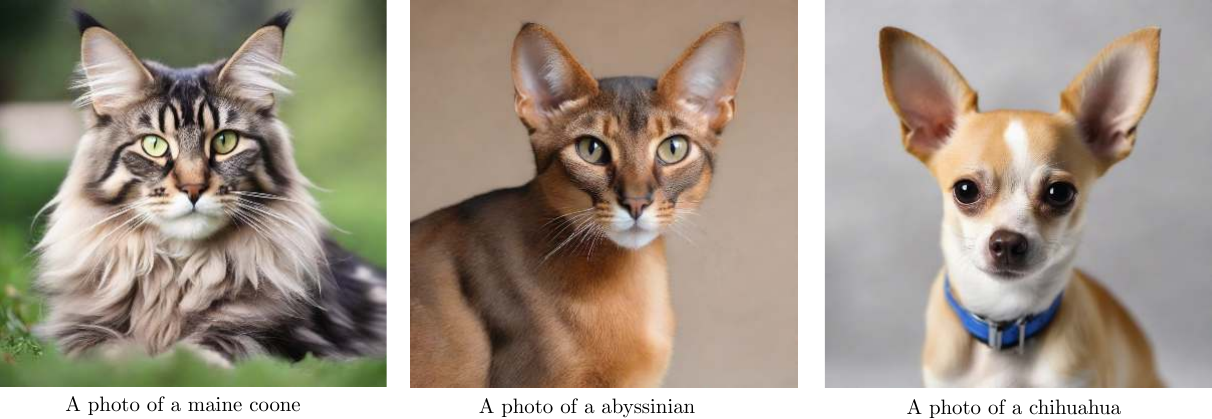}
  \caption{
    Examples from the synthetic dataset set which was generated using the zero-shot prompts "a photo of ...". The resulting images feature less diversity in comparison to the diversified prompts generated by an LLM depicted in Figure \ref{fig:examples}.
  }
  \label{fig:simple_prompts}
\end{figure}

To assess the influence of diverse prompts on image generation, we utilized zero-shot prompts "a photo of a ..., which is a type of ..." to generate images instead of diversified prompts from a LLM. We sample a synthetic pets dataset with the same number of images per class as in the diversified case. Example images are shown in Figure \ref{fig:simple_prompts}. The diversity of the images decreases, especially with regard to the camera angle, as almost all images show only a frontal shot of the animals with the focus on the face. Furthermore, the variety of backgrounds decreases. We fine-tune a TinyViT 11M model on this dataset and observed the results in Table \ref{tab:simple_data}. The accuracy of feature distilled student exhibits only a small decrease in performance, while the models which were fine-tuned with contrastive losses significantly decreased. These findings indicate that the students distilled with the $\mathcal{L}_2$ feature loss can deal better with a lack of diversity during fine-tuning.

\begin{table}[tb]
  \caption{
    Accuracy of the TinyViT 11Ms models pre-trained for one epoch on DataComp medium and fine-tuned on synthetically generated pets test data was sampled with the zero shot prompts "a phot of ..." instead of diverse prompts from a LLM. The performance of feature-distilled students degrade considerably less from a lack of diversity than the contrastively fine-tuned models.
  }
  \label{tab:simple_data}
  \centering
  \resizebox{\textwidth}{!}{
  \begin{tabular}{@{}lcccccc@{}}
    \toprule
    Fine-Tuning Data & Test Data & \hspace*{2.5mm} $\mathcal{L}_2$ \hspace*{3.5mm} & \hspace*{2.5mm} CLIP \hspace*{1.5mm} & MP \hspace*{0.5mm} &  $\mathcal{L}_2$+CLIP \hspace*{1.5mm} & $\mathcal{L}_2$+MP  \\
    \midrule
    Synthetic (simple) & Real  & 85.9 & 40.7 & 45.2 & 81.8 & 49.1 \\
    \textcolor{gray}{Difference to synthetic (diversified)} & & \textcolor{gray}{-1.6} & \textcolor{gray}{-26.0} & \textcolor{gray}{-21.3} & \textcolor{gray}{-5.4} & \textcolor{gray}{-19.0} \\
    Synthetic (simple) \hspace*{1.5mm}  & \hspace*{1.5mm} Synthetic (diversified) & 93.0 & 86.6 & 85.0 & 93.8 & 87.6  \\
    \textcolor{gray}{Difference to synthetic (diversified)} & &\textcolor{gray}{-1.5} & \textcolor{gray}{-10.9} & \textcolor{gray}{-12.7} & \textcolor{gray}{-2.9} & \textcolor{gray}{-10.2}  \\
  \bottomrule
  \end{tabular}}
\end{table}

\subsection{Linear Classification Head Instead Of CLIP Architecture} 
Instead of using the CLIP architecture, we train and distill TinyViT 11M model with a linear classification head on the pets dataset for comparison. We either train from scratch or fine-tune models with pre-trained weights from ImageNet-22k (with the exception of the linear classification head which is always randomly initialized). As most of the classes from the pets dataset are contained in ImageNet-22k, the latter does not correspond to a strict zero-shot setting even when fine-tuning with synthetic data. To train the models, we optimize the standard cross-entropy loss as well as a sum of cross-entropy loss and the original knowledge distillation loss of Hinton et al.\cite{ogdistillation}. We use the AdamW optimizer \cite{adamW} with no weight decay and the learning rate is set to $5\times 10^{-4}$ which is the same as used by Wu et al. \cite{tiny_vit} for fine-tuning. First, we observe that the drop in performance when fine-tuning with synthetic data in comparison to real data is similar to the CLIP models based on contrastive losses. Additionally, the performance of the model with classification head is worse compared to the CLIP models when trained from scratch. In contrast, the classifiers pretrained on ImageNet-22k and fine-tuned on the real training data achieve the best performance overall. This can presumably be attributed to the fact that most of the classes are already included in ImageNet-22k which was used for pre-training. Using knowledge distillation for the models with classification head only has a minor effect.

\begin{table}[tb]
  \caption{
    Accuracy of the models that feature a linear classification head instead of a CLIP architecture. Training was performed on the synthetic and real pets datasets for 96 epochs using the cross-entropy loss (CE) with or without knowledge distillation (KL).
  }
  \label{tab:no_clip}
  \centering
  \begin{tabular}{@{}lcccc@{}}
    \toprule
    Fine-Tuning Data & \hspace*{3.5mm} Pre-Training Data\hspace*{3.5mm}  & \hspace*{1.5mm} CE \hspace*{1.5mm} & \hspace*{1.5mm} CE+KL \hspace*{1.5mm}  \\
    \midrule
    Real & - & 47.8 & 47.2 \\
    Synthetic & - & 26.9 & 29.9 \\
    Real & ImageNet-22k & 91.2 & 91.6 \\
    Synthetic & ImageNet-22k & 71.3 & 67.4 \\
  \bottomrule
  \end{tabular}
\end{table}
\vspace*{16cm}



\end{document}